\documentclass[11pt]{article}

\usepackage[final]{acl}
\usepackage{times}
\usepackage{latexsym}
\usepackage[T1]{fontenc}
\usepackage[utf8]{inputenc}
\usepackage{microtype}
\usepackage{inconsolata}
\usepackage{graphicx}
\usepackage{booktabs}
\usepackage{adjustbox}
\usepackage{rotating}
\usepackage{array}

\usepackage{ifthen}
\usepackage{tikz}
\usetikzlibrary{matrix,calc}
\usepackage{graphicx} 
\usepackage{siunitx} 
\usepackage{tabularx}

\title{ELYADATA \& LIA at NADI 2025: ASR and ADI Subtasks}

\author{
  \textbf{Haroun Elleuch\textsuperscript{1,2,\dag}},
  \textbf{Youssef Saidi\textsuperscript{1,\ddag}},
  \textbf{Salima Mdhaffar\textsuperscript{2}},\\
  \textbf{Yannick Estève\textsuperscript{2}},
  \textbf{Fethi Bougares\textsuperscript{1,2}}\\
  \\
  \textsuperscript{1}ELYADATA, France \quad
  \textsuperscript{2}LIA, Avignon University, France\\
  \\
  \small{\dag~Main contributor of the ADI subtask \quad
  \ddag~Main contributor of the ASR subtask}\\
  \small{\textbf{Correspondence:} \href{mailto:haroun.elleuch@elyadata.com}{haroun.elleuch@elyadata.com}}
}

\begin{document}
\maketitle
\begin{abstract}
This paper describes Elyadata \& LIA's joint submission to the NADI multi-dialectal Arabic Speech Processing 2025. We participated in the Spoken Arabic Dialect Identification (ADI) and multi-dialectal Arabic ASR subtasks. Our submission ranked first for the ADI subtask and second for the multi-dialectal Arabic ASR subtask among all participants. Our ADI system is a fine-tuned Whisper-large-v3 encoder with data augmentation. This system obtained the highest ADI accuracy score of \textbf{79.83\%} on the official test set. For multi-dialectal Arabic ASR, we fine-tuned SeamlessM4T-v2 Large (Egyptian variant) separately for each of the eight considered dialects. Overall, we obtained an average WER and CER of \textbf{38.54\%} and \textbf{14.53\%}, respectively, on the test set. Our results demonstrate the effectiveness of large pre-trained speech models with targeted fine-tuning for Arabic speech processing.
\end{abstract}

\section{Introduction}

Arabic is one of the most widely spoken languages in the world, both in terms of number of speakers and geographical spread \cite{lane2025mostspoken}. This wide distribution, coupled with centuries of contact with other languages and cultures, has led to the emergence of numerous colloquial varieties collectively known as Arabic dialects. Although the exact granularity and classification of these dialects remain a matter of debate, a common working assumption in computational processing is to associate a dialect with a country-level variety \cite{bouamor-etal-2014-multidialectal, shon2020adi17}, or to a larger area where sub-dialects are the most similar (Gulf, Levant, North Africa) \cite{app12178898, ali2017speech}.

Dialectal Arabic poses unique challenges for speech and language processing. Unlike Modern Standard Arabic (MSA), dialects are predominantly spoken rather than written \cite{Ferguson01011959}, with significant variation in phonology, lexicon, and syntax. They also lack standardized orthographic conventions, despite recent efforts such as CODA (Conventional Orthography for Dialectal Arabic) \cite{habash-etal-2012-conventional}, and later efforts of \citet{habash-etal-2018-unified} and \citet{alhafni-etal-2024-exploiting}. These properties complicate both Automatic Speech Recognition (ASR) and Automatic Dialect Identification (ADI) tasks, where systems must generalize across substantial linguistic variability.

The 2025 Nuanced Arabic Dialect Identification (NADI) Shared Task \cite{nadi2025} addresses these challenges through three subtasks aimed at improving the coverage and robustness of speech technologies for Arabic dialects:
\begin{itemize}
    \item Spoken Arabic Dialect Identification (ADI)
    \item Multidialectal Arabic ASR using the recently released Casablanca dataset \cite{talafha2024casablanca}
    \item Diacritic Restoration focusing on dialectal variations of Arabic
\end{itemize}

Our team participated in the first two subtasks, achieving first place in ADI and second place in multi-dialectal ASR on the official test sets. In both cases, we leveraged large-scale pre-trained speech models with targeted fine-tuning strategies to address dialectal variability.

Our main contributions are (1) We propose an effective two-stage fine-tuning approach for ADI, using the Whisper-large-v3 encoder to achieve state-of-the-art results.
(2) We demonstrate that separately fine-tuning the SeamlessM4T-v2 Large model for each dialect yields competitive ASR performances.

\section{Arabic Dialect Identification}
The ADI subtask aims to classify speech utterances into their respective country-level dialect categories automatically. Our approach leverages large-scale pre-trained speech representations and a two-stage fine-tuning process to effectively adapt to the dialectal nuances present in the provided dataset. In the following subsections, we describe the datasets used, our ADI model architecture, and the considered training strategy. We also present our experimental results and follow up with an analysis of the ADI system performances.

\subsection{Datasets}
We utilize several datasets to train and evaluate our ADI system, including established corpora covering multiple Arabic dialects, as well as the official NADI 2025 ADI dataset. The following is a detailed description of each dataset.

\subsubsection{ADI-17 and ADI-20}
The ADI-17 dataset~\cite{shon2020adi17} comprises 3{,}033 hours of dialectal Arabic speech from 17 country-level dialects for training, along with approximately 2 hours per dialect in the development and test splits, respectively.

The ADI-20 dataset~\cite{elleuch2025adi20} is an expanded and rebalanced version of ADI-17, extending its coverage from 17 to 20 Arabic varieties by including Tunisian and Bahraini dialects as well as Modern Standard Arabic (MSA). It also increases representation for previously underrepresented dialects, such as Jordanian and Sudanese, by incorporating additional speech material. In total, the training partition contains 3{,}556 hours of speech, while the development and test sets retain the same structure as in ADI-17, supplemented with approximately 2 hours per newly added variety in each split. To enable experiments under resource-constrained conditions and ensure per-dialect balance, ADI-20-53h, a stratified subset containing up to 53 hours of training data for each variety, resulting in a total of 1{,}060 hours is also available. Our future experiments will use this subset rather than the full ADI-20 dataset for the reasons mentioned earlier.

\subsubsection{NADI 2025 ADI Dataset}
The official dataset for the ADI subtask covers eight country-level Arabic dialects: Algeria, Egypt, Jordan, Mauritania, Morocco, Palestine, UAE, and Yemen.  
It includes an \emph{adaptation} split of approximately 15 hours (12{,}900 utterances) with associated country labels, a validation split of similar size (12{,}700), and an eleven-hour held-out test set with 6268 utterances.

\subsection{Model Architecture}
Our system follows the best-performing configuration from \citet{elleuch2025adi20}. The Whisper-large-v3 encoder~\cite{radford2023robust} is used as a feature extractor, followed by an attention pooling layer that aggregates frame-level representations into fixed-length utterance embeddings. These are passed through a fully connected layer with a softmax activation for classification over the target dialects.

We freeze the first 16 layers of the Whisper encoder during fine-tuning to preserve general speech representations while adapting the upper layers to the ADI task. To enhance robustness, we apply additive noise, speed perturbation, frequency masking, and chunk-level dropout. Training is performed with SpeechBrain~\cite{speechbrainV1} using negative log-likelihood loss, the Adam optimizer, and a NewBob learning rate scheduler starting from $1\times10^{-5}$ for frozen encoder layers and $1\times10^{-4}$ for trainable layers. Training runs for up to 100 epochs on NVIDIA H100 80GB GPUs, with early stopping based on validation performance.

\subsection{Experiments and Results}
We first evaluated the model after fine-tuning only on ADI-17 and ADI-20-53h to assess zero-shot performance on the NADI validation set.  
As shown in Table~\ref{tab:adi_baselines_scores}, fine-tuning on ADI-17 yields an accuracy of 31.84\%, while ADI-20-53h substantially improves zero-shot accuracy to 78.33\%.

\begin{table}[!htb]
\centering
\begin{tabular}{l|c}
\hline
\textbf{Fine-tuning dataset} & \textbf{Accuracy (\%)} \\ \hline
ADI-17 & 31.84 \\
ADI-20-53h & \textbf{78.33} \\
\hline
\end{tabular}
\caption{Zero-shot evaluation on the NADI 2025 ADI validation set.}
\label{tab:adi_baselines_scores}
\end{table}

Our final submission builds on the ADI-20-53h model, further adapted with the NADI adaptation split. This two-stage fine-tuning yields substantial gains, as shown in Table~\ref{tab:adi-results}.  
The system ranked first, achieving 98.08\% accuracy on validation and 79.83\% on the test set, with corresponding average costs of 0.0171 and 0.1788 using the 2022 NIST LRE formulation.\\

Analysis of the validation confusion matrix in~\ref{fig:confmat} shows that the Algerian dialect is the most challenging to predict, with only 96\% of utterances correctly classified. Misclassifications primarily involve the geographically adjacent Moroccan dialect, and conversely, 30 Moroccan utterances are labeled as Algerian. Misclassifications between Egyptian and Jordanian are largely reciprocal; despite their geographic proximity, this pattern is unexpected from the perspective of Arabic speakers.

\begin{figure}[!htb]
    \centering
    \resizebox{1\columnwidth}{!}{\begin{tikzpicture}[
    scale = 1.1, 
    font={\scriptsize}, 
]

\def\myConfMat{{
{1563, 3,    0,    2,    11,   4,    2,    5},
{2,    1579, 9,    0,    0,    3,    0,    4},
{2,    27,   1543, 1,    0,    8,    8,    7},
{7,    0,    0,    1556, 6,    1,    8,    5},
{30,   1,    0,    1,    1549, 1,    5,    5},
{7,    4,    4,    1,    0,    1546, 3,    4},
{2,    4,    3,    3,    0,    0,    1585, 1},
{9,    5,    7,    5,    4,    6,    4,    1535}
}}
\def\classNames{{"ALG","EGY","JOR","MAU","MOR","PAL","UAE","YEM"}}
\def\numClasses{8}

\tikzset{vertical label/.style={rotate=90, anchor=east}}
\tikzset{horizontal label/.style={rotate=0, anchor=center}}

\foreach \y in {1,...,\numClasses} 
{
    \node [anchor=east, font=\normalsize] at (0.4,-\y) {\pgfmathparse{\classNames[\y-1]}\pgfmathresult}; 
    
    \foreach \x in {1,...,\numClasses}  
    {
        \def\totSamples{0}
        \foreach \ll in {1,...,\numClasses}
        {
            \pgfmathparse{\myConfMat[\ll-1][\x-1]}
            \xdef\totSamples{\totSamples+\pgfmathresult}
        }
        \pgfmathparse{\totSamples} \xdef\totSamples{\pgfmathresult}
        
        \begin{scope}[shift={(\x,-\y)}]
            \pgfmathtruncatemacro{\r}{\myConfMat[\y-1][\x-1]}
            \pgfmathtruncatemacro{\p}{round(\r/\totSamples*100)}
            \coordinate (C) at (0,0);
            
            \ifthenelse{\p<50}{\def\txtcol{black}}{\def\txtcol{white}}
            
            \ifnum\x=\y
                \def\celltext{\r\\(\p\%)}
            \else
                \def\celltext{\r}
            \fi
            
            \node[
                draw,
                text=\txtcol,
                align=center,
                fill=black!\p,
                minimum size=1.1*10mm,
                inner sep=0,
                font=\normalsize,
            ] (C) {\celltext};
            
            \ifthenelse{\y=\numClasses}{
            \node [horizontal label, font=\normalsize] at ($(C)-(0,0.9)$)
            {\pgfmathparse{\classNames[\x-1]}\pgfmathresult};}{}
        \end{scope}
    }
}

\coordinate (yaxis) at (-1, 0.5-\numClasses/2);
\coordinate (xaxis) at (0.5+\numClasses/2, -\numClasses-1.5); 
\node [vertical label, font=\normalsize] at (yaxis) {True Label};
\node [font=\normalsize] at (xaxis) {Predicted Label};

\end{tikzpicture}}
    \caption{Confusion matrix on the provided development set.}
    \label{fig:confmat}
\end{figure}

\begin{table}[!htb]
\centering
\begin{tabular}{l|cc}
\hline
\textbf{Split} & \textbf{Accuracy} (\%) $\uparrow$ & \textbf{LRE avg. Cost} $\downarrow$ \\ \hline
Validation & 98.08 & 0.0171 \\
Test       & 79.83 & 0.1788 \\
\hline
\end{tabular}
\caption{Final ADI subtask results.}
\label{tab:adi-results}
\end{table}

\section{Multi-dialectal Arabic ASR}
The multi-dialectal Automatic Speech Recognition (ASR) subtask focuses on transcribing spoken Arabic across eight country-level dialects. This subtask aims to highlight the challenges posed by phonetic, lexical, and syntactic diversity of Arabic dialects. In this section, we describe the dataset, model architecture, training methodology, and the obtained results of our approach.

\subsection{Dataset}
The NADI 2025 ASR dataset includes the same eight dialects as the ADI subtask. Each dialect has 1{,}600 utterances in both the training and validation splits, with durations ranging from 1 to 30 seconds.  
The total duration is 30.72 hours (15.44 hours for training, 15.27 for validation). Table~\ref{tab:asr_dataset_dialects} shows per-dialect durations.

\begin{table}[!htb]
\centering
\begin{tabular}{lcc}
\hline
\textbf{Dialect} & \textbf{Train (h)} & \textbf{Validation (h)} \\ \hline
Algeria     & 1.91 & 1.84 \\
Egypt       & 2.01 & 1.85 \\
Jordan      & 1.93 & 1.89 \\
Mauritania  & 1.66 & 1.63 \\
Morocco     & 1.60 & 1.67 \\
Palestine   & 2.43 & 2.41 \\
UAE         & 1.87 & 1.86 \\
Yemen       & 2.01 & 2.11 \\ \hline 
Total       & \textbf{15.44}    &  \textbf{15.27} \\ \hline
\end{tabular}
\caption{Durations per dialect in the NADI 2025 datasets}
\label{tab:asr_dataset_dialects}
\end{table}

\subsection{Models}
We adopted two distinct architectures in our experiments: Whisper and SeamlessM4T-v2 \cite{barrault2023seamless}.
Whisper is an encoder–decoder Transformer model trained on a large-scale multilingual and multitask dataset of speech and text, enabling robust automatic speech recognition (ASR) across a wide range of languages. Its architecture integrates  a Transformer-based encoder for speech representation learning and a Transformer decoder for transcription generation. The Whisper-large-v3 model contains approximately 1.55 billion parameters, while Whisper-medium has around 769 million parameters, offering a faster and more memory-efficient alternative. \\

SeamlessM4T is a multilingual sequence-to-sequence model designed for speech and text translation across more than 100 languages. In its v2 release, it builds upon the UnitY2 architecture, combining a Conformer-based speech encoder with a Transformer-based text decoder. We selected the Egyptian variant due to its demonstrated effectiveness in Arabic transcription tasks. Given the substantial phonetic, lexical, and syntactic divergence between Arabic dialects, we empirically found that fine-tuning a separate model for each dialect outperformed a single unified model for all dialects. 

\subsection{Experiments and Results}
Our experimental process for the multi-dialectal ASR subtask followed three main steps:  
(i) evaluation of Whisper-based systems,  
(ii) comparison between per-dialect and unified models,  
(iii) comparison of the best Whisper model with SeamlessM4T-v2 Large. All results are reported in terms of WER and CER, computed on the NADI 2025 validation and test sets.\\
For Whisper-based experiments, we fine-tuned different variants (Medium and Large) under two hardware configurations: (i) on an NVIDIA P100 16GB GPU with the AdamW optimizer, a fixed learning rate of $1\times10^{-5}$, a batch size of 1, and gradient accumulation over 4 steps; and (ii) on an NVIDIA A100 80GB GPU with a batch size of 8 using the same optimizer and learning rate.

For SeamlessM4T, we fine-tuned the v2 Large Egyptian variant for six epochs on an NVIDIA A100 40GB GPU. Training employed the AdamW optimizer with a learning rate warmed up over 100 steps from $1\times10^{-9}$ to $5\times10^{-5}$. We used label-smoothed negative log-likelihood loss with a smoothing factor of 0.2 and a batch size of 2.
\subsubsection{Whisper Large vs Whisper Medium}

We first fine-tuned both Whisper-large-v3 and Whisper-medium on the full multi-dialectal dataset (all eight dialects combined).
On average, Whisper-large-v3 achieved a WER of 72.20\% and a CER of 58.51\% while Whisper-medium, despite its smaller size, outperformed it with a WER of 48.21\% and a CER of 17.94\%.
Given these substantial improvements, Whisper-medium was chosen for all subsequent experiments.

\subsubsection{Multi vs Mono-dialectal Models}
We evaluated two training strategies using Whisper-medium: a multi-dialectal model trained jointly on all dialects, and mono-dialectal models obtained via dedicated fine-tuning for each dialect, yielding eight specialized models. The mono-dialectal approach achieved a lower average Word Error Rate (WER) of 46.71\%, compared to 48.21\% for the multi-dialectal model. In terms of Character Error Rate (CER), both approaches performed similarly, with the multi-dialectal model scoring 17.97\% and the specialized models 17.94\% on average. These results suggest that training separate mono-dialectal models is the best approach.

\subsubsection{Whisper vs SeamlessM4T-v2 Large}

We also compared the best Whisper setup (one specialized whisper-medium system per-dialect) with the SeamlessM4T-v2 Large Egyptian model~\cite{barrault2023seamless} also fine-tuned separately for each dialect. Due to time constraints, only the large variant was considered for our experiments.  
Table~\ref{tab:asr_validation_one_model_per_dialect} shows that the Seamless-based system consistently outperforms our best Whisper system for all dialects in both WER and CER.

\begin{table}[!htb]
    \centering
    \small
    \begin{tabular}{l c|c}
        \cline{2-3}
         & \textbf{Seamless} & \textbf{Whisper-med.} \\ \hline
        \multicolumn{1}{c|}{\textbf{Dialect}} &
        \textbf{WER / CER (\%)} &
        \textbf{WER / CER (\%)} \\ \hline
        \multicolumn{1}{l|}{Jordan}     & \textbf{25.26} / \textbf{7.68} & 32.53 / 9.93  \\
        \multicolumn{1}{l|}{Egypt}      & \textbf{30.05} / \textbf{12.52} & 39.38 / 15.97 \\
        \multicolumn{1}{l|}{Morocco}    & \textbf{39.24} / \textbf{13.48} & 49.22 / 18.34 \\
        \multicolumn{1}{l|}{Algeria}    & \textbf{54.13} / \textbf{19.34} & 60.61 / 22.41 \\
        \multicolumn{1}{l|}{Yemen}      & \textbf{50.49} / \textbf{16.85} & 61.28 / 25.54 \\
        \multicolumn{1}{l|}{Mauritania} & \textbf{56.93} / \textbf{23.91} & 62.79 / 26.97 \\
        \multicolumn{1}{l|}{UAE}        & \textbf{30.90} / \textbf{10.59} & 35.38 / 12.48 \\
        \multicolumn{1}{l|}{Palestine}  & \textbf{26.35} / \textbf{9.64} & 32.51 / 12.11 \\
        \hline
        \multicolumn{1}{l|}{Average}    & \textbf{39.17} / \textbf{14.25} & 46.71 / 17.97 \\
        \hline
    \end{tabular}
    \caption{SeamlessM4T-v2 Large vs. Whisper-medium WER and CER on the validation sets of each NADI 2025 dialect using one fine-tuned model per dialect.}
    \label{tab:asr_validation_one_model_per_dialect}
\end{table}

\subsubsection{Official Submission }
Based on the validation results presented in table~\ref{tab:asr_validation_one_model_per_dialect}, we selected the SeamlessM4T-v2 Large per-dialect models for submission.  
It ranked second overall, with an average WER 38.54\% and CER 14.53\% on the test set. Table~\ref{tab:asr_results_test} shows the results per dialect. As it can be seen, performance varied notably across dialects, with Levantine and Egyptian achieving the lowest WERs, while Maghrebi dialects remained the most challenging.

\begin{table}[!htb]
\centering
\begin{tabular}{lcc}
\hline
\textbf{Dialect} & \textbf{WER (\%)} & \textbf{CER (\%)} \\ \hline
Jordan     &  28.03 & 9.36  \\
Egypt      & 26.83 & 11.44 \\
Morocco    & 38.27 & 13.66 \\
Algeria    & 53.73 & 20.43 \\
Yemen      & 46.63 & 16.66 \\
Mauritania & 58.11 & 24.53 \\
UAE        & 29.35 & 9.91  \\
Palestine  & 27.36 & 10.20 \\ \hline
\end{tabular}
\caption{WER and CER on the NADI 25 test sets.}
\label{tab:asr_results_test}
\end{table}

\section{Conclusion}
This paper presented the ELYADATA–LIA submissions to the NADI 2025 shared task, addressing both the Arabic Dialect Identification and Multidialectal Automatic Speech Recognition subtasks.
For ADI, we demonstrated the effectiveness of a two-stage fine-tuning approach using the Whisper-large-v3 encoder, achieving first place with 79.83\% accuracy on the test set.
For ASR, fine-tuning the SeamlessM4T-v2 Large model separately for each dialect resulted in a strong performance, ranking second on the leaderboard with an average WER of 38.54\%.

\section*{Acknowledgments}
This work was partially funded by the ESPERANTO project. The ESPERANTO project has received funding from the European Union’s Horizon 2020 (H2020) research and innovation program under the Marie Skłodowska-Curie grant agreement No 101007666.
This work was granted access to the HPC resources of IDRIS under the allocations AD011015051R1 and AD011012551R3 made by GENCI.

\bibliography{custom}

\end{document}